%% file: main.tex
\documentclass[10pt,twocolumn,letterpaper]{article}

\usepackage[pagenumbers]{cvpr} 
\input{preamble}

\definecolor{cvprblue}{rgb}{0.21,0.49,0.74}
\usepackage[pagebackref,breaklinks,colorlinks,allcolors=cvprblue]{hyperref}
\usepackage{amsmath,amssymb,bm}
\newcommand{\argmin}{\operatorname*{arg\,min}}

\usepackage{booktabs}
\usepackage{multirow}
\usepackage{graphicx}

\usepackage{xcolor}

\def\confName{CVPR}
\def\confYear{2027}

\title{InitFree-SfM: \\[0.3em] When Initialization-Free Bundle Adjustment \\ Meets Learned Structure-from-Motion \vspace{-2mm}}

\author{Simon Weber\textsuperscript{1} \hspace{2em}
Mateo de Mayo\textsuperscript{2,3} \hspace{2em}
Je Hyeong Hong\textsuperscript{4}  \hspace{2em}
Carl Olsson\textsuperscript{5}  \\
Daniel Cremers\textsuperscript{2,3}\hspace{2em} 
Ronald Clark\textsuperscript{1}  \hspace{2em}
\\[3mm]
\textsuperscript{1}University of Oxford \hspace{1cm} \textsuperscript{2}Technical University of Munich \hspace{1cm} \textsuperscript{3}Munich Center for Machine Learning \\
\textsuperscript{4}Hanyang University \hspace{1cm}
\textsuperscript{5}Lund University 
\vspace{-3mm}
}

\begin{document}
\maketitle
%\vspace{-15mm}
\input{sec/0_abstract}
\vspace{-5mm}
\input{sec/1_intro}
\input{sec/2_related_work}
\input{sec/4_method}
\input{sec/5_experiments}

\input{sec/6_conclusion}

{
    \small
    \bibliographystyle{ieeenat_fullname}
    \bibliography{main}
}

 \input{sec/X_suppl}

\end{document}

%% file: preamble.tex
\usepackage{cuted} %< for strip environment

\usepackage{currfile} %< for \currfilebase macro

\usepackage{caption} %< for "\captionof" commands (teaser)
\makeatletter
\renewcommand\paragraph{\@startsection{paragraph}{4}{\z@}{0.6em}{-1em}{\normalfont\normalsize\bfseries}}
\makeatother

\usepackage{tikz}
\usetikzlibrary{positioning, arrows.meta, calc, fit, backgrounds, decorations.pathreplacing, shapes.geometric}
\definecolor{FigGreen}{HTML}{004225}
\definecolor{FigGreenPastel}{HTML}{E0EEE5}
\definecolor{FigRose}{HTML}{8B3433}
\definecolor{FigRosePastel}{HTML}{F4DEDD}
\definecolor{FigOchre}{HTML}{8A6A1F}
\definecolor{FigOchrePastel}{HTML}{FBF1DC}
\definecolor{FigSand}{HTML}{D8D2C6}
\definecolor{FigSandPastel}{HTML}{F7F5EE}
\definecolor{FigInk}{HTML}{2F302C}

\usepackage{booktabs}
\usepackage{multirow}
\usepackage{arydshln}
\usepackage{array}
\newlength{\tablestdmeanheight}
\newlength{\tablestdsmallheight}

\usepackage{amsthm}
\usepackage{placeins}

%% file: sec/0_abstract.tex
\begin{abstract}
Initialization-free bundle adjustment (InitFree BA) aims to jointly recover camera poses and scene structure without requiring an initial global estimate. It is a very challenging problem, especially compared to traditional BA where a good estimation is assumed. While recent methods optimize surrogate Object Space Error (OSE) objectives from random initialization, a low objective value does not necessarily yield an accurate reconstruction, revealing a gap between optimization and reconstruction. We show that existing InitFree BA pipelines struggle to produce accurate 3D reconstructions on modern SfM benchmarks, and demonstrate that incorporating learned metric depth substantially improves their practical feasibility. We introduce InitFree-SfM\footnote{Project page: \url{https://github.com/simonwebertum/initfree-sfm.git}}, a complete initialization-free SfM pipeline combining learned correspondences, metric depth, depth-consistent track construction, and joint metric refinement. Within this framework, we propose a simple depth-augmented OSE (dOSE), which complements the directional constraints of OSE by constraining landmark positions along their viewing rays. Our approach substantially outperforms existing InitFree BA methods and achieves competitive accuracy with established SfM pipelines, including on challenging low-overlap image collections. Importantly, our analysis challenges the prevailing emphasis on surrogate objectives in previous InitFree BA research, revealing that geometric constraints play a central role in accurate reconstruction, while OSE-based optimization provides further improvements.
\end{abstract}

%% file: sec/1_intro.tex
\begin{figure*}[t]
\centering
\includegraphics[width=.98\textwidth]
{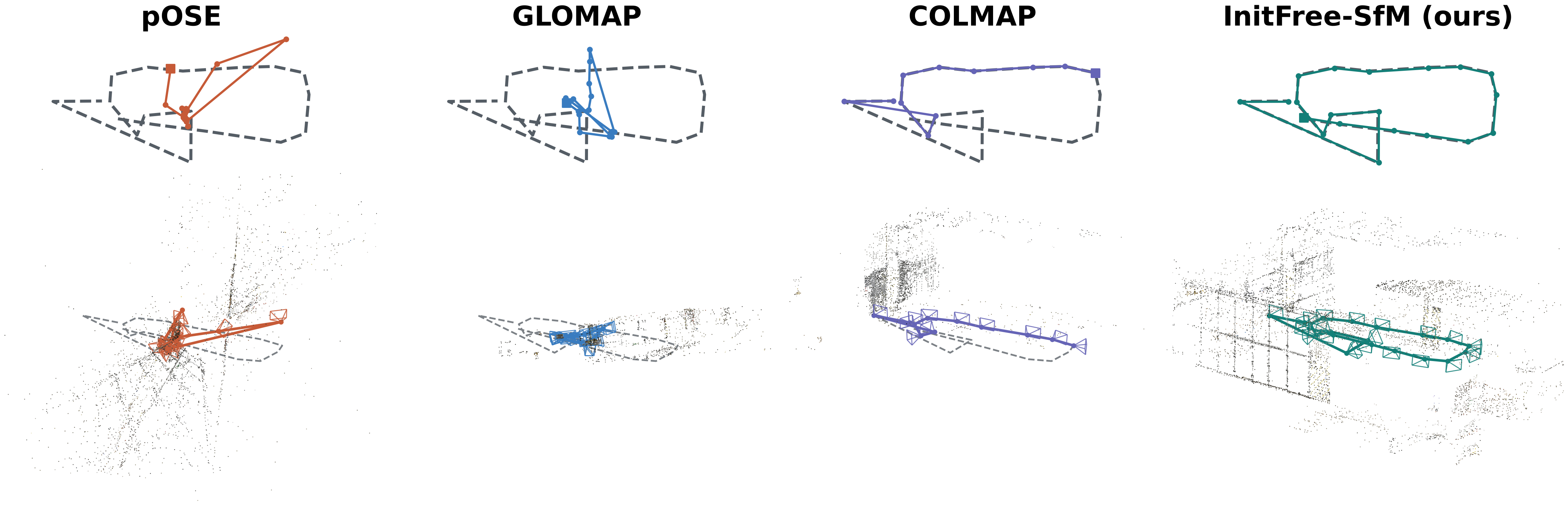}
\par\vspace{-1mm}
{\small\textbf{(a)} Sparse SfM comparison: delivery area}
\par\vspace{6mm}
\includegraphics[width=.98\textwidth]
{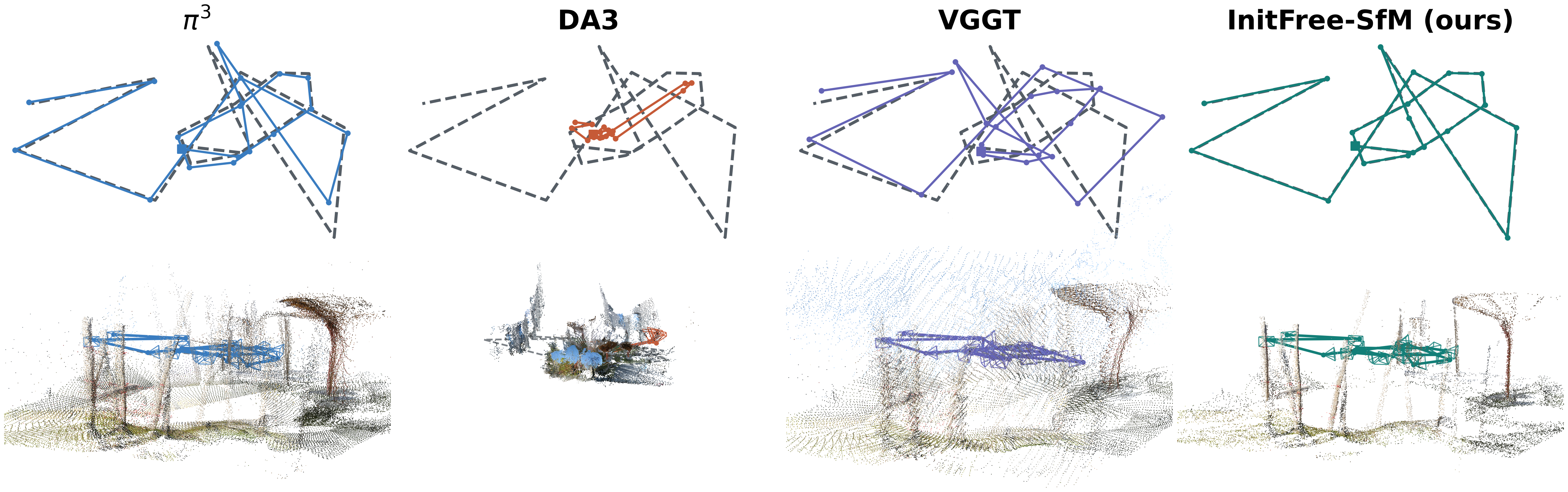}
\par\vspace{-1mm}
{\small\textbf{(b)} Comparison with feed-forward methods: playground}
\caption{
\textbf{Qualitative comparisons on ETH3D \cite{schops2017multi} using the same 20 images per scene.} \textbf{(a)} pOSE (LoMa), GLOMAP (LoMa), COLMAP (SP+LG) and Ours (LoMa), with joint pose AUC@$5^\circ$ of 0.00, 37.36, 28.00, and 87.34\%, respectively. \textbf{(b)} $\pi^3$, DA3, VGGT and Ours (RoMaV2) scoring 43.38, 34.25, 10.83, and 95.05\%, respectively. Gray dashed lines connect ground-truth camera centers.}
\label{fig:eth3d-sparse-dense}
\end{figure*}

\begin{figure*}[t]
  \centering
  \includegraphics[width=\textwidth]{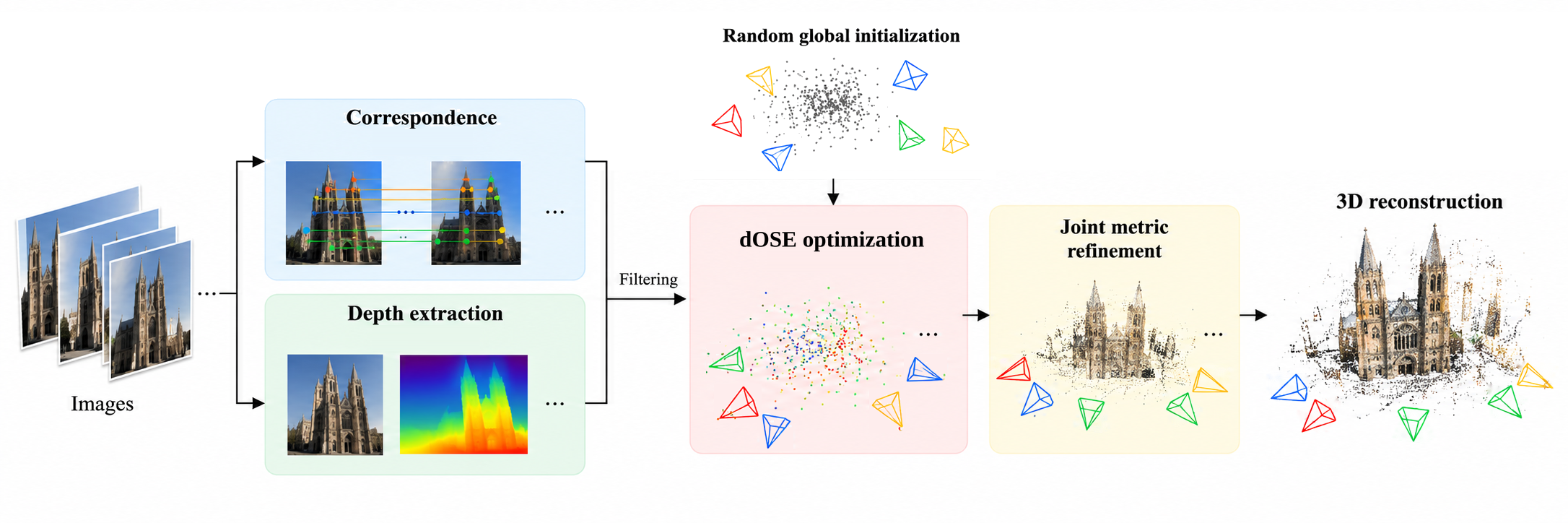}
  \caption{\textbf{Overview of InitFree-SfM.} Starting from input images, we extract pairwise correspondences and monocular metric depth, followed by geometric consistency filtering (\cref{subsec:depth_estimation,subsec:depth_consistent_correspondences}). From random global initialization, dOSE jointly optimizes camera poses and 3D landmarks (\cref{subsec:depth_based_OSE}), followed by joint metric refinement incorporating depth and relative pose constraints to obtain the final 3D reconstruction (\cref{subsec:joint_metric_refinement}). 
Overall, our simple yet effective pipeline offers a new perspective on initialization-free BA, highlighting the central role of geometric constraints beyond surrogate objective optimization.
  }
  \label{fig:pipeline}
\end{figure*}

\section{Introduction}
\label{sec:intro}

The joint optimization of camera poses and scene landmarks, known as bundle adjustment (BA), is a central component of structure-from-motion (SfM) pipelines \cite{hartley2003multiple}. In its traditional formulation \cite{triggs1999bundle}, BA is used as a refinement stage, after suitable initial estimates of the camera poses and scene structure have been obtained, typically through a sequence of geometric steps such as relative pose estimation, rotation averaging, and triangulation.

A recent line of work \cite{zach2018pose,iglesias2021radial,iglesias2023expose,olsson2025towards} challenges this paradigm by considering BA with random initialization. This setting, referred to as \textit{initialization-free bundle adjustment} (InitFree BA), seeks to jointly optimize global camera poses and scene landmarks without requiring an initial estimate of either. To tackle this challenging problem, existing methods typically adopt a three-stage pipeline: \textit{(i)} a surrogate Object Space Error (OSE), combined with regularization, is optimized using variable projection (VarPro); \textit{(ii)} camera parameters and landmarks are refined by minimizing reprojection error; and \textit{(iii)} a metric upgrade \cite{pollefeys1999self} is applied to recover Euclidean camera poses from relaxed camera matrices in $\mathbb{R}^{3\times4}$. 

Despite this promising optimization framework, recent experimental evidence \cite{weber2026initialization} reveals a substantial gap between optimization and reconstruction: existing OSE-based pipelines can converge to low surrogate objective values while producing geometrically inaccurate reconstructions. This discrepancy arises from insufficient geometric constraints during optimization and the instability of the subsequent metric upgrade. Consequently, successful surrogate optimization does not necessarily translate into accurate reconstruction, as will be further illustrated in App.~\ref{app:gap_convergence}.

At the same time, recent advances in learned geometry offer an opportunity to revisit this problem. Monocular depth foundation models provide geometric information that complements classical multi-view constraints. While OSE encourages reconstructed landmarks to lie along their observed viewing rays, predicted depth additionally constrains their positions along these rays. More generally, learned geometric priors provide information that can be exploited throughout the reconstruction pipeline, rather than only during its final optimization. This raises a natural question: \textit{Can learned geometric priors bridge the gap between optimization and reconstruction in initialization-free BA?}

In this work, we investigate this question by introducing InitFree-SfM, a complete structure-from-motion pipeline built around initialization-free joint optimization. We integrate learned correspondences and monocular metric depth through depth-consistent correspondence filtering, initialization-free OSE-based optimization, and joint metric refinement. In particular, we revisit the classical OSE formulation by introducing a simple depth-augmented objective (dOSE), and directly optimize camera poses on $SE(3)$ using variable projection, avoiding the need for a subsequent metric upgrade. Finally, we jointly refine the reconstruction using image observations, predicted depths, and pairwise geometric constraints.

Our experiments demonstrate that the resulting pipeline substantially improves reconstruction accuracy over existing InitFree BA approaches, whose conventional optimization and metric-upgrade procedures often fail on modern datasets. Moreover, InitFree-SfM achieves competitive accuracy with established SfM pipelines, including on challenging low-overlap image collections. Surprisingly, we find that joint metric refinement can already recover accurate reconstructions directly from random initialization, without any intermediate OSE optimization. This challenges the prevailing emphasis on surrogate objectives in previous InitFree BA research, suggesting that the availability of geometric constraints is the primary driver of accurate reconstruction, while OSE-based initialization provides additional improvements.

\begin{itemize}

    \item \textit{Revisiting initialization-free BA.}
    We demonstrate that existing initialization-free BA formulations often fail to produce accurate reconstructions on modern SfM benchmarks. We show that incorporating learned depth information throughout the reconstruction pipeline substantially improves their practical applicability.

    \item \textit{A complete initialization-free SfM pipeline.}
    We introduce InitFree-SfM, combining learned correspondences, monocular metric depth, depth-consistent correspondence filtering, initialization-free joint optimization directly on $SE(3)$, and joint metric refinement. We additionally propose a simple depth-augmented Object Space Error (dOSE) that retains the variable-projection structure of existing formulations.

    \item \textit{New insights into initialization-free reconstruction.}
    Contrary to the emphasis on surrogate objective design in previous work, we show that accurate reconstruction can be achieved through joint metric refinement directly from random initialization, without an OSE stage. Our experiments highlight the central role of geometric constraints, while demonstrating that dOSE further improves reconstruction accuracy.

    \item \textit{Competitive reconstruction accuracy.}
    We demonstrate substantial improvements over existing InitFree BA approaches and competitive performance against modern SfM pipelines on ETH3D, LaMAR, and challenging low-overlap benchmarks.

\end{itemize}

We hope that our novel formulation and promising experimental results will open new perspectives on the challenging and still relatively unexplored problem of initialization-free bundle adjustment.

%% file: sec/2_related_work.tex
\section{Related Work}
\label{sec:related_work}

As our approach bridges feasible structure-from-motion to initialization-free bundle adjustment, we review works on recent SfM pipelines and on InitFree BA.

\paragraph{Traditional and feed-forward structure-from-motion}

Traditional SfM pipelines recover camera poses and scene structure through a sequence of geometric operations, including feature matching, relative pose estimation, triangulation, and bundle adjustment \cite{hartley2003multiple}. Incremental approaches, such as COLMAP \cite{schoenberger2016sfm}, progressively register images and reconstruct landmarks, while global approaches, such as GLOMAP~\cite{pan2024global}, estimate camera poses jointly from pairwise geometric constraints. Recent advances in learned geometry bridge SfM pipelines to learned feature matching and monocular geometric priors. Methods such as VGGSfM \cite{wang2024vggsfm} integrate learned components into incremental reconstruction, while MP-SfM~\cite{pataki2025mpsfm} exploits monocular surface priors to improve robustness under limited image overlap. Recently, feed-forward approaches such as Dust3r~\cite{wang2024dust3r}, VGGT~\cite{wang2025vggt}, $\pi^3$ \cite{wang2026pi}, and Depth Anything 3 \cite{lin2025depth} directly predict multi-view geometry from images through a single pass. Despite these advances, recovering accurate 3D reconstructions by jointly optimizing camera poses and scene structure directly from random initialization remains largely unexplored.

\paragraph{Initialization-free bundle adjustment.}
InitFree BA aims to jointly recover global camera poses and scene structure without requiring an initial estimate of their global configuration (see App.~\ref{app:definition} for a formal definition). The problem, linked to the photogrammetry community \cite{fusiello2015solving}, is relatively novel, and highly challenging. Early work demonstrated that projective BA could be optimized from arbitrary initialization using variable projection algorithm~\cite{hong2016projective}, exploiting the separable structure of the Object Space Error (OSE). Subsequent approaches, including pOSE~\cite{zach2018pose}, RpOSE \cite{iglesias2021radial}, expOSE~\cite{iglesias2023expose}, pOSE+rot~\cite{olsson2025towards}, improve the optimization of OSE-based objectives through different regularizations. Weber et al~\cite{weber2024power} extend the InitFree BA problem to large-scale datasets. However, these methods have primarily focused on the optimization problem itself, rather than its integration into a complete SfM pipeline. Moreover, recent experimental study~\cite{weber2026initialization} highlights a substantial discrepancy between minimizing surrogate OSE objectives and recovering accurate reconstructions, particularly due to the subsequent metric upgrade. 
Despite their theoretical appeal, the potential of InitFree BA for practical SfM remains largely unexplored. Our work revisits this problem from a different perspective, emphasizing the role of geometric constraints in enabling accurate metric reconstruction through initialization-free joint optimization.

%% file: sec/4_method.tex
\section{Depth-constrained Object Space Error}
Let us first formalize the object space error formulation (\cref{subsec:OSE}), that is the cornerstone of InitFree BA approaches \cite{zach2018pose}, and our proposed depth-based OSE (\cref{subsec:depth_based_OSE}).

\subsection{Background: Object Space Error}\label{subsec:OSE}

Let $\tilde X_j = (X_j^\top,1)^\top$ denote the homogeneous
coordinates of the $j$-th 3D point. Under the pinhole camera model,
its observation $m_{ij}^{\mathrm{pixel}}=(u_{ij},v_{ij})^\top$ in
image $i$ satisfies
\begin{equation}
    \lambda_{ij}
    \begin{pmatrix}
        u_{ij} \\
        v_{ij} \\
        1
    \end{pmatrix}
    =
    K_i
    \begin{pmatrix}
        R_i & t_i
    \end{pmatrix}
    \tilde X_j ,
\end{equation}
where $\lambda_{ij}$ is the projective depth.
We assume the cameras are calibrated, and subsequently we work throughout with normalized
image coordinates. For each pixel observation $m_{ij}^{\mathrm{pix}}=(u_{ij},v_{ij})^\top$, we define the corresponding normalized homogeneous viewing ray as:
\begin{equation}
    \tilde m_{ij}
    =
    K_i^{-1}
    \begin{pmatrix}
        u_{ij}\\v_{ij}\\1
    \end{pmatrix},
    \qquad
    \tilde m_{ij,3}=1,
    \qquad
    m_{ij}=\tilde m_{ij,1:2}.
\end{equation}
Absorbing the calibration into the camera matrix gives
$P_i=[R_i\;t_i]$. Eliminating the unknown projective depth by using the third coordinate of $P_i\tilde X_j$ yields the object-space residual
\begin{equation}
    r_{ij}^{\mathrm{OSE}}
    =
    P_{i,1:2}\tilde X_j
    -
    (p_{i,3}^{\top}\tilde X_j)m_{ij},
\end{equation}
where $P_{i,1:2}$ denotes the first two rows of $P_i$ and
$p_{i,3}^{\top}$ its third row. The corresponding OSE objective is
\begin{equation}
    \mathcal{L}_{\mathrm{OSE}}
    =
    \sum_{(i,j)\in\mathcal O}
    \left\|
        r_{ij}^{\mathrm{OSE}}
    \right\|_2^2,
\end{equation}
where $\mathcal O$ denotes the set of observations.
The OSE residual admits a simple geometric interpretation: it measures whether the transformed landmark lies on the viewing ray associated with its image observation. Indeed, the residual vanishes exactly when $P_i\tilde X_j$ is collinear with the normalized ray $\tilde m_{ij}$. 

\subsection{Depth-based Object Space Error}\label{subsec:depth_based_OSE}

However, OSE is only a directional constraint and does not constrain
the position of each transformed landmark along its observed viewing
ray. We complement this directional constraint with metric depth
predictions. Let $d_{ij}$ denote the predicted depth at observation
$m_{ij}$, and let
\begin{equation}
    y_{ij} = R_i X_j + t_i
\end{equation}
be landmark $j$ expressed in the coordinate system of camera $i$.
The scalar
\begin{equation}
    q_{ij}^{\top}y_{ij} \equiv \frac{\tilde m_{ij}^{\top}}
         {\tilde m_{ij}^{\top}\tilde m_{ij}} y_{ij}
\end{equation}
is the orthogonal projection coefficient of $y_{ij}$ onto the observed ray $\tilde m_{ij}$, and therefore represents its depth along this ray. Indeed, whenever $y_{ij}$ lies on the ray, i.e., $y_{ij}=z\tilde m_{ij}$ for some depth $z$, the above expression is equal to $z$. Based on that, we augment the classical OSE objective with a robust depth-consistency term, yielding our depth-based Object Space Error (dOSE):
\begin{align}\label{eq:dOSE}
    \mathcal{L}_{\mathrm{dOSE}}
    = 
    &\sum_{(i,j) \in \mathcal{O}}
    c_{ij}
    \left\|
        P_{i,1:2}\tilde X_j
        -
        (p_{i,3}^{\top}\tilde X_j)m_{ij}
    \right\|_2^2
    \nonumber\\
    &+
    \lambda_d
    \sum_{(i,j) \in \mathcal{O}}
    c_{ij}
    \rho\left(
        \frac{ q_{ij}^{\top}y_{ij}}
        {d_{ij}}
        - 1
    \right),
\end{align}
where $c_{ij}$ denotes the confidence associated with observation $(i,j)$, $\rho$ is the Huber loss, and $\lambda_d$ controls the contribution of the depth term. The normalization by $d_{ij}$ makes the depth residual relative rather than absolute, preventing large-depth observations from dominating the objective.

The two terms provide complementary geometric constraints: the OSE term constrains the direction of $y_{ij}$ by encouraging it to lie on the observed viewing ray, while the depth term constrains its position along that ray to agree with the predicted metric depth. Consequently, the objective encourages the reconstructed point toward $y_{ij}=d_{ij}\tilde m_{ij}$ in camera coordinates. Our dOSE formulation (\Cref{eq:dOSE}) retains the separable structure of classical OSE objectives \cite{zach2018pose,iglesias2023expose}, allowing us to eliminate the landmarks using variable projection \cite{hong2017revisiting}.

\paragraph{Variable Projection.}
 Let $u=\{R_i,t_i\}_{i=1}^{M-1}$ denote the camera poses, with the first camera fixed to remove the global gauge, and $v=\{X_j\}_{j=1}^{N}$ the landmarks. The objective can be written as
\begin{equation}
    \min_{u,v} \varepsilon(u,v)
    = \|G(u)v-z(u)\|_2^2,
    \qquad (R_i,t_i)\in SE(3).
\end{equation}
For fixed camera poses, the optimal landmarks are eliminated in closed form as $v^*(u)=G(u)^\dagger z(u)$, reducing the nonlinear optimization to the camera poses alone. Unlike previous OSE formulations \cite{zach2018pose,weber2023power,iglesias2023expose}, which optimize relaxed camera matrices in $\mathbb{R}^{3\times4}$, our depth-augmented formulation directly optimizes metric poses on $SE(3)$, avoiding the unstable metric upgrade \cite{weber2026initialization}.

We solve the reduced problem with the Ruhe--Wedin (RW2) approximation \cite{kaufman1975variable,hong2017revisiting}. For small problems, we use Cholesky factorization; for larger one, we employ the inverse-expansion solver of Weber et al. \cite{weber2023power,weber2024power}. Further details are provided in App.~\ref{app:varpro}.

\begin{figure*}[t]
  \centering
  \includegraphics[width=\textwidth]{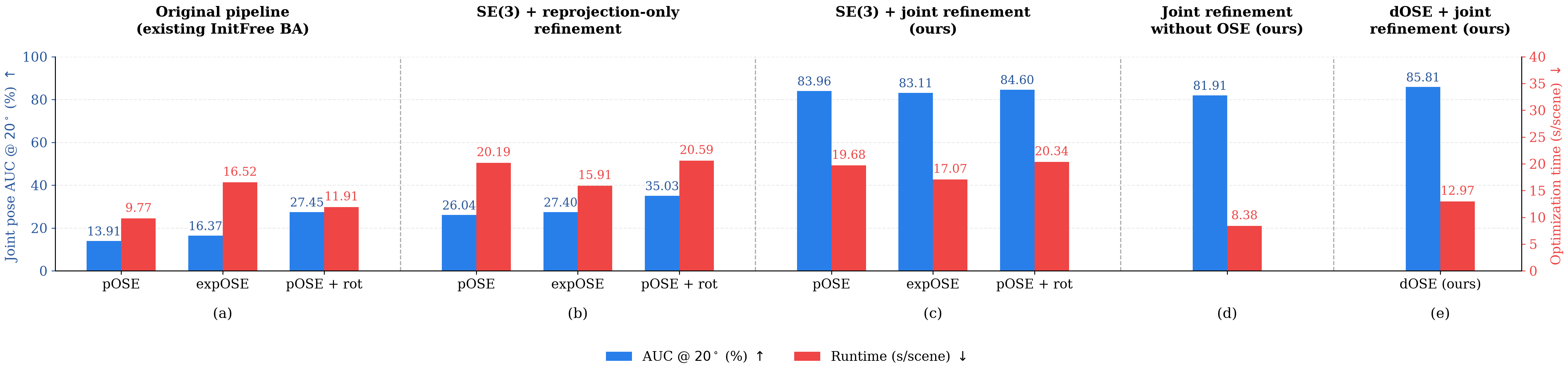}
  \caption{\textbf{Dissecting initialization-free BA on ETH3D.}
We compare (a) traditional OSE-based pipelines, (b) OSE optimization on $SE(3)$ with reprojection refinement, (c) OSE optimization with our joint metric refinement, (d) joint metric refinement directly from random initialization, and (e) our complete dOSE-based pipeline. Geometrically constrained refinement yields substantial improvements, even without preceding OSE optimization, while dOSE achieves the best accuracy. All configurations use the same correspondences and are randomly initialized (see App. \ref{app:settings}). Bars report pose AUC@$20^\circ$ and mean optimization time per scene. }
  \label{fig:initfree_practical}
\end{figure*}

\section{Initialization-Free SfM Pipeline}
\label{sec:pipeline}

We now integrate dOSE into a complete structure-from-motion pipeline (\cref{fig:pipeline}), combining learned correspondences and monocular metric depth (\cref{subsec:depth_estimation}), geometric correspondence filtering (\cref{subsec:depth_consistent_correspondences}), initialization-free optimization, and joint metric refinement (\cref{subsec:joint_metric_refinement}). Unlike previous works focusing on InitFree BA as a standalone optimization problem \cite{iglesias2023expose,olsson2025towards,iglesias2021radial}, we investigate its integration into a complete SfM pipeline.

\subsection{Depth Estimation}\label{subsec:depth_estimation}

We first extract metric geometric information independently from each image using MoGe-2 \cite{wang2026moge}. For every image $i$, MoGe-2 predicts a metric depth map $d_i$, a validity mask, and normalized pinhole intrinsics, which we can convert to the original image resolution to obtain the camera calibration matrix $K_i$ (see \cref{subsec:ablation}) for uncalibrated cases.

We bilinearly sample the predicted depth map at the observation location $m_{ij}^{\mathrm{pix}}=(u_{ij},v_{ij})^\top$ to obtain $d_{ij}=d_i(m_{ij}^{\mathrm{pix}})$. The observation can then be back-projected into the local coordinate system of camera $i$ as
\begin{equation}
    p_{ij}^{\mathrm{M}}
    =
    d_{ij}\tilde m_{ij}.
\end{equation}
These local metric points provide both the depth cues used in our dOSE formulation and an additional geometric signal for validating image correspondences, as described next.

\subsection{Depth-Consistent Correspondences}
\label{subsec:depth_consistent_correspondences}

Given pairwise image correspondences, obtained for instance using LoMa \cite{nordstrom2026loma}, we exploit predicted metric geometry to identify consistent image pairs and correspondences. For each image pair $(k,l)$, we back-project the correspondences using predicted depth and robustly align the resulting 3D point sets using RANSAC to estimate a similarity transformation $(s_{lk}^{*},R_{lk}^{*},t_{lk}^{*})$:
\begin{align}
    \argmin_{\substack{s>0,\,R\in SO(3),\\t\in\mathbb{R}^3}}
    \sum_{j\in\mathcal{C}_{kl}}
    \rho\left(
        \left\|
            p_{lj}^{\mathrm{M}}
            -
            \left(sRp_{kj}^{\mathrm{M}}+t\right)
        \right\|_2
    \right),
\end{align}
where $\mathcal{C}_{kl}$ denotes the correspondences between images $k$ and $l$. 

Since predicted depth can be unreliable, we cross-validate the estimated transformation against calibrated epipolar geometry. From the normalized homogeneous rays $\tilde m_{kj}$ and $\tilde m_{lj}$, we independently estimate an essential matrix satisfying
\begin{equation}
    \tilde m_{lj}^{\top}E_{lk}\tilde m_{kj}=0,
    \qquad
    E_{lk}=[\tilde t_{lk}]_{\times}\tilde R_{lk},
    \label{eq:essential_crosscheck}
\end{equation}
yielding a relative rotation $\tilde R_{lk}$ and translation direction $\tilde t_{lk}$. We retain only image pairs whose depth-based and epipolar rotations agree within a threshold $\tau_R$.

For accepted pairs, we combine the rotation and translation direction from epipolar geometry with the translation magnitude and relative scale from depth alignment:
\begin{equation}
    \bar R_{lk}=\tilde R_{lk},
    \qquad
    \bar t_{lk}
    =
    \lVert t_{lk}^{*}\rVert_2
    \frac{\tilde t_{lk}}
         {\lVert\tilde t_{lk}\rVert_2},
    \qquad
    \bar s_{lk}=s_{lk}^{*}.
    \label{eq:cross_validated_pose}
\end{equation}
Finally, the surviving correspondences are merged into multi-view tracks for subsequent initialization-free BA. Importantly, these pairwise geometric estimates are used solely as optimization constraints, without constructing an initial global camera configuration or 3D reconstruction.

\subsection{Joint Metric Refinement}
\label{subsec:joint_metric_refinement}

Existing OSE-based approaches typically perform reprojection-error refinement \cite{zach2018pose,iglesias2023expose}, sometimes followed by post-processing to enforce geometric validity. However, reprojection refinement alone may be insufficient to recover accurate reconstructions \cite{weber2026initialization}. We instead propose a joint metric refinement that exploits the predicted depths and relative geometric constraints already available in our pipeline, alongside image observations.

We jointly optimize camera poses and landmarks, introducing a per-image depth scale $\alpha_i$ to account for inconsistencies \cite{yu2025relative} in monocular depth predictions.
For each observation $(i,j)$ and focal length $(f_x,f_y)$, we define the normalized image residuals, equivalent to the classical reprojection error scaled by the ratio between reconstructed and predicted depths, and depth residuals:
\begin{align}
    r_{ij}^{\text{img}}
        &= \frac{1}{\alpha_i d_{ij}}
        \begin{pmatrix}
            f_x(y_{ij,1}-m_{ij,1}y_{ij,3})\\
            f_y(y_{ij,2}-m_{ij,2}y_{ij,3})
        \end{pmatrix}, \\
    r_{ij}^{\text{depth}}
        &= \frac{q_{ij}^{\top}y_{ij}-\alpha_i d_{ij}}
        {\alpha_i d_{ij}}.
\end{align}
For each accepted image pair $(k,l)\in\mathcal{E}$, we enforce consistency with the relative pose and scale estimates $(\bar R_{kl},\bar t_{kl},\bar s_{kl})$ from \Cref{eq:cross_validated_pose}. We scale relative translations by the corresponding depth scales to ensure consistency with the global camera translations and we use logarithmic residuals to penalize multiplicative scale errors symmetrically:
\begin{align}
    r_{kl}^{\text{rot}}
        &= R_kR_l^\top-\bar R_{kl},\\
    r_{kl}^{\text{t}}
        &= \frac{t_k-R_kR_l^\top t_l-\alpha_k\bar t_{kl}}
        {\alpha_k\lVert \bar t_{kl} \rVert_2},\\
    r_{kl}^{\text{s}}
        &= \frac{\log\alpha_l-\log\alpha_k-\log\bar s_{kl}}
        {\sigma_s}.
\end{align}

We first constrain the metric translation using $r_{kl}^{\text{t}}$, before switching to a translation-direction constraint,
\begin{equation}
    r_{kl}^{\text{t,dir}}
    =
    \frac{t_k-R_kR_l^\top t_l}
         {\|t_k-R_kR_l^\top t_l\|_2}
    -
    \frac{\bar t_{kl}}{\|\bar t_{kl}\|_2},
\end{equation}
making the refinement less sensitive to errors in the translation magnitude predicted from monocular depth (see App.~\ref{app:warm_up} for details).
The complete refinement objective combines these terms with cheirality and scale regularization, and is solved with the Levenberg-Marquardt algorithm:
\begin{align}
    \mathcal{L}_{\text{ref}}
    ={}& E_{\text{img}}+E_{\text{depth}}
       +E_{\text{cheirality}}+E_{\text{scale}} \nonumber\\
      &+\sum_{(k,l)\in\mathcal E}\|r_{kl}^{\text{rot}}\|_2^2
       +\sum_{(k,l)\in\mathcal E}\|r_{kl}^{\text{t}}\|_2^2
       +\sum_{(k,l)\in\mathcal E}\rho(r_{kl}^{\text{s}}).
\end{align}

%% file: sec/5_experiments.tex
\section{Experiments}\label{sec:experiments}

We revisit existing InitFree BA pipelines, demonstrating how learned depth and geometric constraints substantially improve geometric reconstruction (\cref{subsec:initfree_ba_sota}). We then evaluate InitFree-SfM against established SfM approaches, including on challenging low-overlap datasets (\cref{subsec:exp_established,subsec:low_overlap}). Finally, we conduct ablation studies to assess the contributions of individual components (\cref{subsec:ablation}). We report pose AUC@1°/5°/20°, following \cite{pan2024global,wang2025vggt}. For InitFree BA methods, camera rotations are sampled uniformly on $\mathrm{SO}(3)$, translations from a Gaussian distribution, and landmarks are recovered through VarPro. We give more details, as well as dataset and implementation settings, in App.~\ref{app:settings}. 

\subsection{Can InitFree BA recover accurate reconstructions?}\label{subsec:initfree_ba_sota}

Following common practice \cite{wang2025vggt}, we evaluate existing OSE-based InitFree BA methods on 25 ETH3D scenes \cite{schops2017multi}, with up to 20 images each. We consider pOSE~\cite{zach2018pose}, expOSE~\cite{iglesias2023expose}, and pOSE+rot~\cite{olsson2025towards}, progressively examining different optimization and refinement strategies (\cref{fig:initfree_practical}). The original pipelines, combining relaxed camera optimization in $\mathbb{R}^{3\times4}$, reprojection refinement, and metric upgrade, yield poor accuracy (Fig.~3a). Direct optimization on $SE(3)$ improves accuracy but remains insufficient with reprojection-only refinement (Fig.~3b). Introducing our depth-aware joint metric refinement substantially improves all three OSE formulations (Fig.~3c). Surprisingly, this refinement alone recovers accurate reconstructions directly from random initialization, without any OSE optimization (Fig.~3d). This suggests that geometric constraints during refinement, rather than surrogate OSE optimization alone, are central to accurate reconstruction. Nevertheless, our proposed dOSE further improves accuracy, achieving the best results in this comparison (Fig.~3e).
\vspace{-5mm}

\subsection{Can InitFree-SfM compete with modern SfM pipelines?}\label{subsec:exp_established}

\begin{table*}[t]
\centering
\footnotesize
\setlength{\tabcolsep}{3.5pt}
\renewcommand{\arraystretch}{1.05}
\caption{\textbf{Comparison with established SfM methods.}
Pose AUC (\%) on LaMAR \cite{sarlin2022lamar} and ETH3D \cite{schops2017multi}. Bold and underlined values denote the best and second-best results. To our knowledge, InitFree-SfM is the first SfM method built around InitFree BA that is competitive with modern SfM pipelines, substantially outperforming prevailing InitFree BA pipeline pOSE.}
\label{tab:initfree_competitive}
\begin{tabular}{@{}clrrrrrr@{}}
\toprule
& \multirow{2}{*}{Method}
& \multicolumn{3}{c}{LaMAR}
& \multicolumn{3}{c}{ETH3D} \\
\cmidrule(lr){3-5}\cmidrule(lr){6-8}
&
& AUC@$1^\circ$ & AUC@$5^\circ$ & AUC@$20^\circ$
& AUC@$1^\circ$ & AUC@$5^\circ$ & AUC@$20^\circ$ \\
\midrule

\multirow{2}{*}{\rotatebox[origin=c]{90}{\scriptsize Incr.}}
& VGGSfM \cite{wang2024vggsfm}
& 10.18 & 42.62 & 57.14
& 19.61 & 42.33 & 53.78 \\
& COLMAP \cite{schoenberger2016sfm} (SP+LG)
& 24.72 & 54.33 & 65.68
& \underline{57.88} & 68.56 & 72.66 \\
\midrule

\multirow{4}{*}{\rotatebox[origin=c]{90}{\scriptsize Feed-forward}}
& VGGT \cite{wang2025vggt}
& 3.85 & 40.55 & 69.95
& 3.59 & 31.16 & 62.99 \\
& VGGT + BA \cite{wang2025vggt}
& 5.67 & 44.09 & 70.44
& 7.13 & 38.67 & 68.03 \\
& $\pi^3$ \cite{wang2026pi}
& 8.47 & 54.06 & 79.48
& 7.56 & 42.33 & 73.96 \\
& DA3 \cite{lin2025depth}
& 5.76 & 48.52 & \underline{79.71}
& 8.36 & 44.93 & 71.56 \\
\midrule

\multirow{4}{*}{\rotatebox[origin=c]{90}{\scriptsize Global}}
& GLOMAP \cite{pan2024global} (SP+LG)
& \textbf{28.02} & 56.76 & 66.87
& 55.30 & 63.71 & 67.20 \\
& GLOMAP \cite{pan2024global} (LoMa)
& \underline{27.28} & \underline{57.45} & 67.93
& \textbf{64.50} & \textbf{73.48} & \underline{77.18} \\
\cmidrule(lr){2-8}
& pOSE \cite{zach2018pose}
& 0.07 & 2.83 & 10.10
& 9.10 & 11.43 & 13.91 \\
& InitFree-SfM (LoMa) (ours)
& 20.23 & \textbf{59.61} & \textbf{80.99}
& 44.40 & \underline{72.63} & \textbf{85.81} \\
\bottomrule
\end{tabular}
\end{table*}

We evaluate InitFree-SfM on ETH3D \cite{schops2017multi} and LaMAR \cite{sarlin2022lamar}, comparing against incremental (COLMAP, VGGSfM), feed-forward (VGGT, $\pi^3$, DA3), and global (GLOMAP) SfM methods, as well as pOSE \cite{zach2018pose}. As shown in \cref{tab:initfree_competitive}, InitFree-SfM achieves the highest AUC@$20^\circ$ on both ETH3D and LaMAR, demonstrating strong overall reconstruction accuracy. While traditional SfM methods remain more accurate at the strictest $1^\circ$ threshold, our results highlight the potential of initialization-free joint optimization to recover competitive accuracy.

\subsection{How robust is InitFree-SfM to reduced image overlap?}\label{subsec:low_overlap}

\begin{figure}[t]
    \centering
    \includegraphics[width=0.49\linewidth]{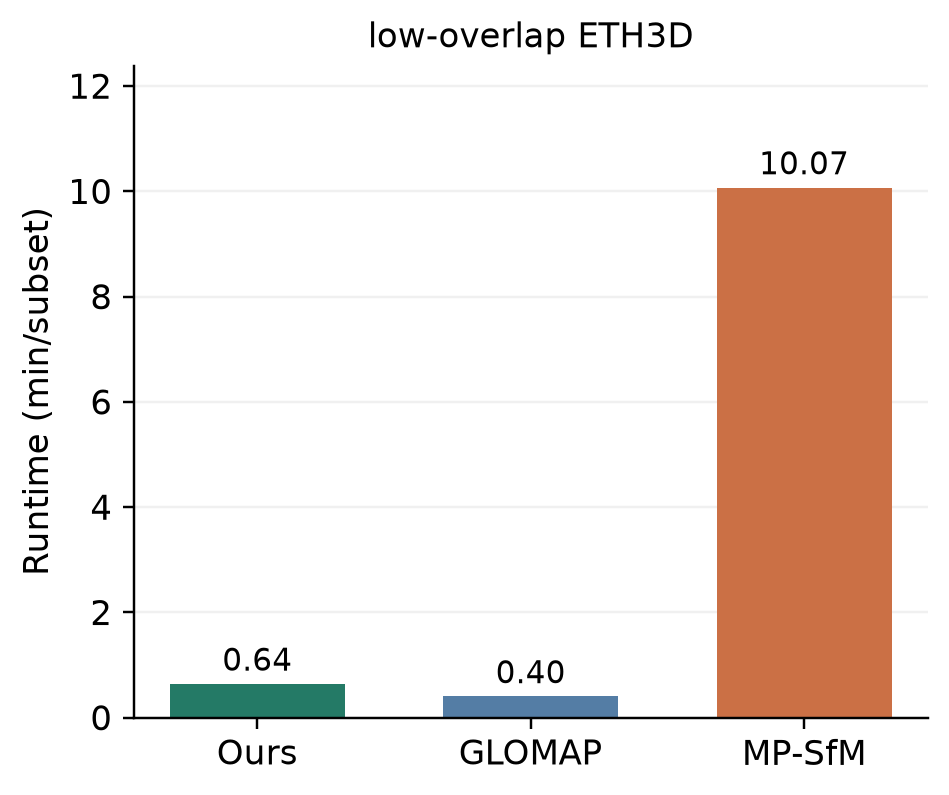}
    \hfill
    \includegraphics[width=0.49\linewidth]{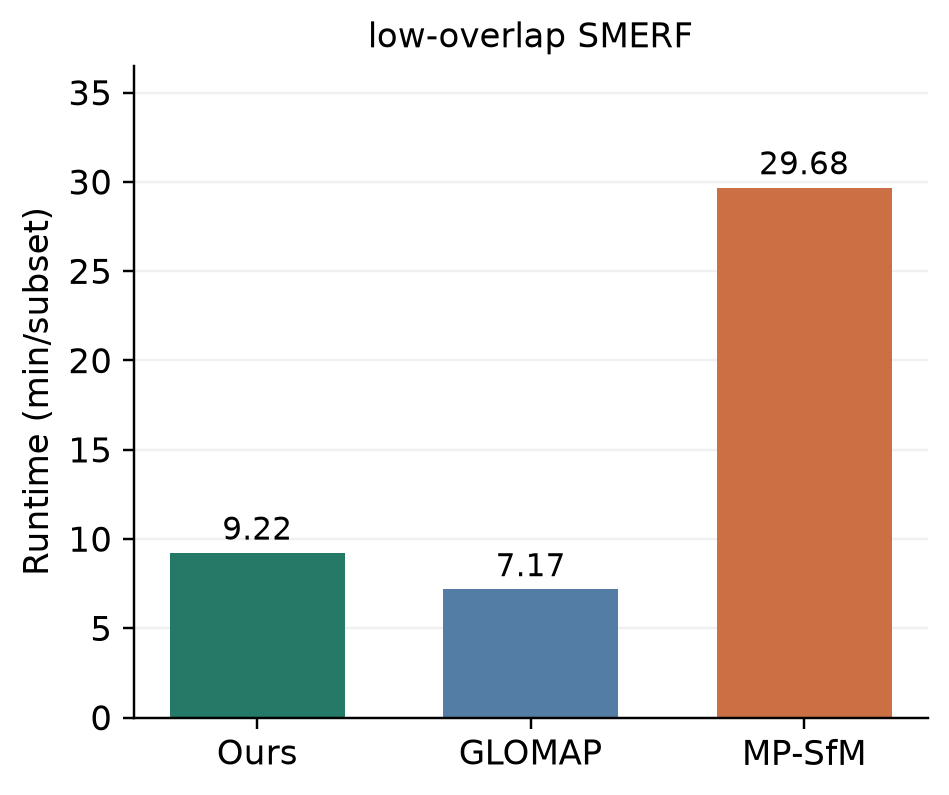}

    \caption{\textbf{Runtime comparison on low-overlap ETH3D (left) and SMERF (right).} Lower is better. InitFree-SfM offers a favorable accuracy-runtime trade-off, achieving competitive reconstruction accuracy at substantially lower computational cost than MP-SfM. All methods use LoMa; runtimes are averaged per subset.}
    \label{fig:runtime-low-overlap}
    \vspace{-5mm}
\end{figure}

We evaluate InitFree-SfM on low-overlap subsets of ETH3D \cite{schops2017multi} and SMERF \cite{duckworth2024smerf}, following the protocol of \cite{pataki2025mpsfm}. We consider both sparse (LoMa \cite{nordstrom2026loma}) and dense (RoMa v2 \cite{edstedt2026roma}) matchers, and include MP-SfM \cite{pataki2025mpsfm}, specifically designed for low-overlap reconstruction, among our baselines. As shown on \cref{tab:low-overlap}, on ETH3D, InitFree-SfM achieves the second-best accuracy in 23 out of 24 comparisons across overlap levels, accuracy thresholds, and matchers, while requiring substantially less computation than the leading MP-SfM method (\cref{fig:runtime-low-overlap}). On SMERF, InitFree-SfM also achieves the second-best accuracy in 10 out of 24 comparisons, although the performance gap to MP-SfM is larger, particularly with sparse correspondences. This coincides with more fragmented graphs of verified relative pose constraints (54\% for SMERF against 68\% for ETH3D low-overlap after depth-checking in the filtering with LoMa). These results show that InitFree-SfM remains competitive under challenging low-overlap conditions, offering a favorable accuracy--runtime trade-off, particularly on ETH3D.

%\vspace{-7mm}

\begin{table*}[t]
\centering
\scriptsize
\setlength{\tabcolsep}{3pt}
\renewcommand{\arraystretch}{1.15}
\caption{\textbf{Low-overlap reconstruction on ETH3D and SMERF}, following~\cite{pataki2025mpsfm}. InitFree-SfM achieves competitive accuracy, particularly on ETH3D, while requiring significantly shorter runtimes than MP-SfM (see \cref{fig:runtime-low-overlap}). Each cell reports AUC@1$^\circ$/5$^\circ$/20$^\circ$ (\%). Within each matcher group, bold and underlining denote the best and second-best results at each threshold. OOM denotes an out-of-memory failure.}
\label{tab:low-overlap}
\resizebox{\textwidth}{!}{%
\begin{tabular}{@{}clcccccccc@{}}
\toprule
& \multirow{2}{*}{Method}
& \multicolumn{4}{c}{ETH3D: 434 subsets from 19 scenes}
& \multicolumn{4}{c}{SMERF: 100 subsets from four scenes} \\
\cmidrule(lr){3-6}\cmidrule(lr){7-10}
&
& Minimal (167) & $\leq5\%$ (84)
& $\leq10\%$ (94) & $\leq30\%$ (89)
& Minimal (40) & Low (20) & Medium (20) & High (20) \\
\midrule

\multirow{4}{*}{\rotatebox[origin=c]{90}{Other SfM}}
& COLMAP \cite{schoenberger2016sfm} (SP+LG)
& 7.9/12.7/14.6 & 6.9/12.5/15.4
& 12.0/19.8/23.2 & 44.9/60.3/65.0
& 2.2/4.2/4.9 & 3.4/7.8/9.9
& 17.7/30.7/36.0 & 42.9/55.4/59.5 \\

& VGGSfM \cite{wang2024vggsfm}
& 0.9/3.2/5.1 & 0.6/2.7/5.0
& 1.8/5.5/8.6 & 7.8/19.6/25.7
& 0.0/0.1/0.3 & 0.0/0.0/0.3
& OOM & OOM \\

& GLOMAP \cite{pan2024global} (SP+LG)
& 8.4/15.8/22.5 & 5.5/15.3/25.7
& 12.1/25.3/35.7 & 50.1/66.7/71.8
& 3.5/8.0/13.3 & 3.4/9.6/15.9
& 20.9/35.3/42.8 & 48.1/60.9/65.4 \\

& MASt3R-SfM \cite{duisterhof2025mast3r}
& 20.1/39.7/52.2 & 17.2/37.5/49.2
& 19.8/37.3/48.1 & 31.4/50.4/59.2
& 3.9/10.4/18.0 & 4.3/11.7/23.0 & 5.9/15.8/28.1 & 10.4/22.9/39.9 \\

\midrule

\multirow{4}{*}{\rotatebox[origin=c]{90}{Feed-forward}}
& VGGT \cite{wang2025vggt}
& 0.72/15.62/47.77 & 0.74/16.07/48.36
& 1.15/18.84/50.22 & 1.84/22.84/56.00
& 0.04/3.10/20.14 & 0.05/4.07/24.65
& 0.03/4.25/26.40 & 0.06/4.83/25.40 \\

& VGGT + BA \cite{wang2025vggt}
& 1.52/16.53/46.44 & 1.44/19.02/51.11
& 2.35/22.50/54.17 & 3.39/27.62/59.58
& 0.07/4.26/22.77 & 0.09/6.01/29.28
& 0.07/6.36/31.28 & OOM \\

& $\pi^3$ \cite{wang2026pi}
& 2.26/27.68/61.38 & 2.42/30.61/63.84
& 2.47/29.82/63.57 & 4.51/36.12/68.05
& 0.17/13.36/47.40 & 0.11/10.53/44.95
& 0.15/13.16/48.95 & 0.20/14.06/47.51 \\

& DA3 \cite{lin2025depth}
& 3.08/30.85/62.72 & 2.92/30.07/62.93
& 2.96/32.90/65.41 & 5.90/42.05/73.07
& 0.01/4.34/25.55 & 0.01/3.71/27.43
& 0.02/5.13/34.25 & 0.02/4.90/30.43 \\

\midrule

\multirow{4}{*}{\rotatebox[origin=c]{90}{LoMa (s)}}
& GLOMAP \cite{pan2024global}
& 13.33/20.37/26.56
& 10.71/21.09/29.69
& 23.10/36.45/44.57
& \textbf{55.51}/66.69/70.66
& \underline{6.30}/12.44/16.94
& \textbf{36.03}/\textbf{48.64}/\underline{54.86}
& \textbf{51.04}/\textbf{64.13}/\underline{69.22}
& \textbf{63.90}/\textbf{74.37}/\underline{77.95} \\

& MP-SfM \cite{pataki2025mpsfm}
& \textbf{28.28}/\textbf{59.26}/\textbf{76.55}
& \textbf{26.81}/\textbf{60.06}/\textbf{78.70}
& \textbf{27.67}/\textbf{61.24}/\textbf{78.46}
& \underline{43.30}/\textbf{73.48}/\textbf{86.70}
& \textbf{11.08}/\textbf{49.92}/\textbf{78.02}
& \underline{9.48}/\underline{42.29}/\textbf{70.30}
& \underline{13.76}/\underline{51.43}/\textbf{79.77}
& \underline{24.72}/\underline{60.66}/\textbf{85.78} \\

& pOSE \cite{zach2018pose}
& 0.00/0.10/0.93 & 0.00/0.27/1.03
& 0.00/0.03/0.89 & 0.00/0.05/0.89
& 0.00/0.00/0.07 & 0.00/0.00/0.03
& 0.00/0.00/0.04 & 0.00/0.00/0.03 \\

& InitFree-SfM (ours)
& \underline{26.88}/\underline{58.03}/\underline{75.31}
& \underline{23.24}/\underline{55.18}/\underline{72.80}
& \underline{25.10}/\underline{56.76}/\underline{73.51}
& 36.17/\underline{66.82}/\underline{81.97}
& 5.69/\underline{27.92}/\underline{45.16}
& 8.14/31.79/52.30
& 13.01/38.89/56.02
& 11.91/35.51/49.12 \\

\midrule

\multirow{4}{*}{\rotatebox[origin=c]{90}{RoMa v2 (d)}}
& GLOMAP \cite{pan2024global}
& 0.82/3.11/7.15 & 0.50/3.05/7.23
& 1.94/6.06/12.99 & 9.05/20.37/33.43
& 0.06/0.59/2.38 & 7.11/20.99/30.61
& \textbf{23.27}/36.74/42.76
& \textbf{39.66}/\textbf{51.19}/\underline{57.06} \\

& MP-SfM \cite{pataki2025mpsfm}
& \textbf{45.14}/\textbf{72.04}/\textbf{87.96}
& \textbf{46.74}/\textbf{73.92}/\textbf{87.37}
& \textbf{49.98}/\textbf{77.19}/\textbf{89.69}
& \textbf{65.11}/\textbf{86.67}/\textbf{94.87}
& \textbf{11.23}/\textbf{47.10}/\textbf{74.18}
& \textbf{9.40}/\textbf{39.77}/\textbf{67.49}
& \underline{12.40}/\textbf{44.88}/\textbf{76.03}
& \underline{19.31}/\underline{50.31}/\textbf{77.38} \\

& pOSE \cite{zach2018pose}
& 0.00/0.04/0.61 & 0.00/0.14/0.86
& 0.00/0.03/0.69 & 0.00/0.01/0.44
& 0.00/0.00/0.03 & 0.00/0.00/0.02
& 0.00/0.00/0.02 & 0.00/0.00/0.04 \\

& InitFree-SfM (ours)
& \underline{32.41}/\underline{60.59}/\underline{78.06}
& \underline{30.20}/\underline{59.80}/\underline{76.66}
& \underline{33.84}/\underline{64.52}/\underline{80.98}
& \underline{38.82}/\underline{69.96}/\underline{86.35}
& \underline{6.21}/\underline{27.46}/\underline{44.38}
& \underline{8.31}/\underline{34.33}/\underline{48.46}
& 10.18/\underline{37.84}/\underline{54.00}
& 13.64/39.70/54.65 \\

\bottomrule
\end{tabular}%
}
\par\smallskip
\raggedright

\end{table*}

\begin{table}[t]
\centering
\small
\setlength{\tabcolsep}{5pt}
\renewcommand{\arraystretch}{1.15}
\caption{\textbf{Robustness to the matcher.} InitFree-SfM, here with MoGe-2, achieves competitive accuracy across sparse and dense matching strategies, with RoMa v2 performing best on ETH3D and MASt3R on LaMAR. Pose AUC@1$^\circ$/5$^\circ$/20$^\circ$ (\%).}
\label{tab:ablation-matching}
\begin{tabular}{lcc}
\toprule
Correspondences & ETH3D & LaMAR \\
\midrule
LoMa (2048) \cite{nordstrom2026loma}
& 44.40/72.63/85.81 & 20.23/59.61/80.99 \\
LoMa (4096) \cite{nordstrom2026loma}
& 49.76/77.17/88.37 & 22.35/59.33/79.12 \\
MASt3R \cite{leroy2024grounding}
& 36.03/69.27/83.05 & 25.47/63.73/83.86 \\
SP + LG \cite{lindenberger2023lightglue}
& 31.20/58.35/69.44 & 16.45/53.85/74.29 \\
RoMa v2 \cite{edstedt2026roma}
& 50.26/77.46/89.33 & 24.08/61.91/80.33 \\
%RoMa V2 (2048) \cite{edstedt2026roma}
%& 49.94/77.77/89.25 & 23.70/61.30/79.60 \\
\bottomrule
\end{tabular}
\end{table}

\begin{table}[t]
\centering
\small
\setlength{\tabcolsep}{7pt}
\renewcommand{\arraystretch}{1.15}
\caption{\textbf{Robustness to the depth estimator.}
InitFree-SfM, here with LoMa, achieves consistent reconstruction accuracy across different monocular depth foundation models. Pose AUC@1$^\circ$/5$^\circ$/20$^\circ$ (\%).}
\label{tab:ablation-depth}
\begin{tabular}{lcc}
\toprule
Depth model
& ETH3D 
& LaMAR  \\
\midrule
MoGe-2 \cite{wang2026moge} & 44.40/72.63/85.81 & 20.23/59.61/80.99 \\
MoGe-3 \cite{kong2026moge} & 43.58/71.59/83.24 & 20.80/59.27/79.66 \\
DA3 \cite{lin2025depth}   & 42.41/71.01/83.84 & 19.86/57.50/77.62 \\
\bottomrule
\end{tabular}
\vspace{-3mm}
\end{table}

\begin{table}[]
\centering
\small
\setlength{\tabcolsep}{5pt}
\renewcommand{\arraystretch}{1.15}
\caption{\textbf{Ablation of InitFree-SfM.} Joint metric refinement is essential for accurate reconstruction, while dOSE provides further improvements. Pose AUC@1$^\circ$/5$^\circ$/20$^\circ$ (\%). Depth and geometric priors refer to constraints in joint refinement.}
\label{tab:ablation-components}
\begin{tabular}{lcc}
\toprule
Configuration & ETH3D & LaMAR \\
\midrule
Ours (full)
& 44.40/72.63/85.81 & 20.23/59.61/80.99 \\
w/o depth filtering
& 32.25/60.04/75.51 & 19.35/56.24/74.57 \\
w/o depth priors
& 32.75/61.95/78.84 & 15.64/51.35/72.65 \\
w/o geometric priors
& 49.69/69.63/78.73 & 21.47/55.98/74.29 \\
w/o refinement
& 0.11/5.65/31.99 & 1.89/16.97/46.48 \\
w/o OSE
& 41.81/68.82/81.91 & 19.77/55.26/73.70 \\
\bottomrule
\end{tabular}
\end{table}

\subsection{Ablation study}\label{subsec:ablation}

\paragraph{Robustness to the matcher.}
\cref{tab:ablation-matching} demonstrates the robustness of InitFree-SfM across a wide range of correspondence extractors, from sparse to dense matchers. While RoMa v2 achieves the best results on ETH3D and MASt3R on LaMAR, our pipeline remains competitive across different matching strategies.

\paragraph{Robustness to the depth estimator.}
\cref{tab:ablation-depth} evaluates the impact of different monocular depth foundation models. The results remain consistent across MoGe-2, MoGe-3, and DA3. This suggests that our approach does not rely on a particular depth estimator.
\vspace{-5mm}

\paragraph{Contribution of individual components.}
As shown in \cref{tab:ablation-components}, depth-consistent filtering and geometric priors improve reconstruction accuracy. 
Joint metric refinement is particularly important: removing it causes a substantial performance drop. Although refinement alone achieves competitive accuracy from random initialization, initializing it with dOSE further improves the results. Additional low-overlap experiments in App.~\ref{app:random_joint} confirm this benefit, with larger gains on SMERF.

\paragraph{Uncalibrated approach.}

\begin{figure}[t]
    \centering
    \includegraphics[width=0.95\linewidth]
    {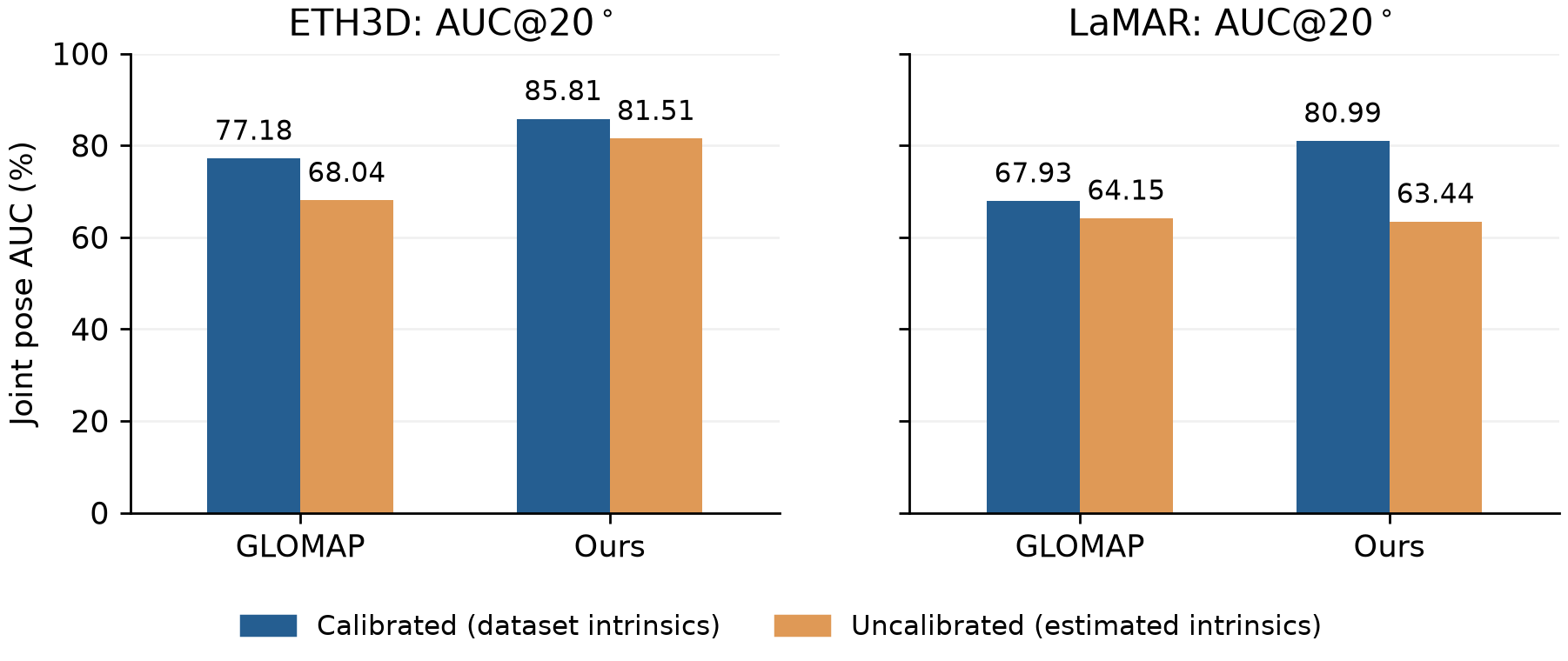}
    \caption{\textbf{Effect of calibration on ETH3D and LaMAR}. Calibrated runs use dataset intrinsics; uncalibrated runs use MoGe-2 intrinsics for ours and native calibration estimation for GLOMAP.}
    \label{fig:calibration}
    \vspace{-3mm}
\end{figure}

Our method remains competitive without dataset calibration (\cref{fig:calibration}): on ETH3D, estimated intrinsics retain 81.51\% AUC@20°, exceeding even calibrated GLOMAP (77.18\%). On LaMAR, accuracy reaches 63.44\%, comparable to uncalibrated GLOMAP (64.15\%) and approaching VGGT (69.95\%). However, the accuracy drop relative to using known intrinsics, particularly on LaMAR, highlights the sensitivity of our method to calibration errors.

%% file: sec/6_conclusion.tex
\section{Conclusion}\label{sec:conclusion}

We revisit initialization-free BA by investigating how recent advances in learned metric geometry can enable accurate 3D reconstruction from random initialization. Building on this insight, we introduce InitFree-SfM, a complete SfM pipeline combining learned metric geometry with a depth-augmented Object Space Error, recovering camera poses and scene structure without a subsequent metric upgrade. Our analysis highlights the central role of geometric constraints in accurate reconstruction, alongside the benefits of dOSE optimization. Experiments demonstrate substantial improvements over prevailing InitFree BA methods and competitive accuracy with modern SfM pipelines, even under low overlap. Overall, our findings offer a promising perspective on InitFree BA, hopefully encouraging further research into this challenging, largely unexplored problem.

\paragraph{Limitations and future work.}
Our method benefits from known camera calibration, motivating improvements in uncalibrated reconstruction. Scaling to larger image collections and exploring additional geometric priors remain promising directions.

%\clearpage

\section{LLM usage statement}
LLMs were not used to generate research ideas and content of the related work. ChatGPT-6 Luna was used to polish the manuscript (grammar and writing) after an initial draft entirely written by the authors, generate code following our instructions, aggregate experimental results into tables (Tab. 1--5), generate bar plots (Fig. 3--5) and edit the visualization in Fig. 2.

%% file: sec/X_suppl.tex
\clearpage
\setcounter{page}{1}
\maketitlesupplementary
\appendix

\noindent This supplemental material is organized as follows:\\
\noindent \textbf{App.~\ref{app:gap_convergence}} illustrates the gap between OSE convergence and metric reconstruction accuracy.\\
\noindent \textbf{App.~\ref{app:definition}} formalizes the definition of initialization-free BA.\\
\noindent \textbf{App.~\ref{app:random_joint}} evaluates the contribution of dOSE on low-overlap datasets.\\
\noindent \textbf{App.~\ref{app:depth_filtering}} investigates the effect of the depth filtering.\\
\noindent \textbf{App.~\ref{app:warm_up}} gives more details about our joint metric refinement.\\
\noindent \textbf{App.~\ref{app:settings}} provides initialization and implementation details.\\
\noindent \textbf{App.~\ref{app:varpro}} details the variable projection algorithm. \\
\noindent \textbf{App.~\ref{app:sophisticated}} investigates additional OSE formulations with relative rotation constraints.

\begin{figure*}[t]
    \centering
    \includegraphics[width=\linewidth]{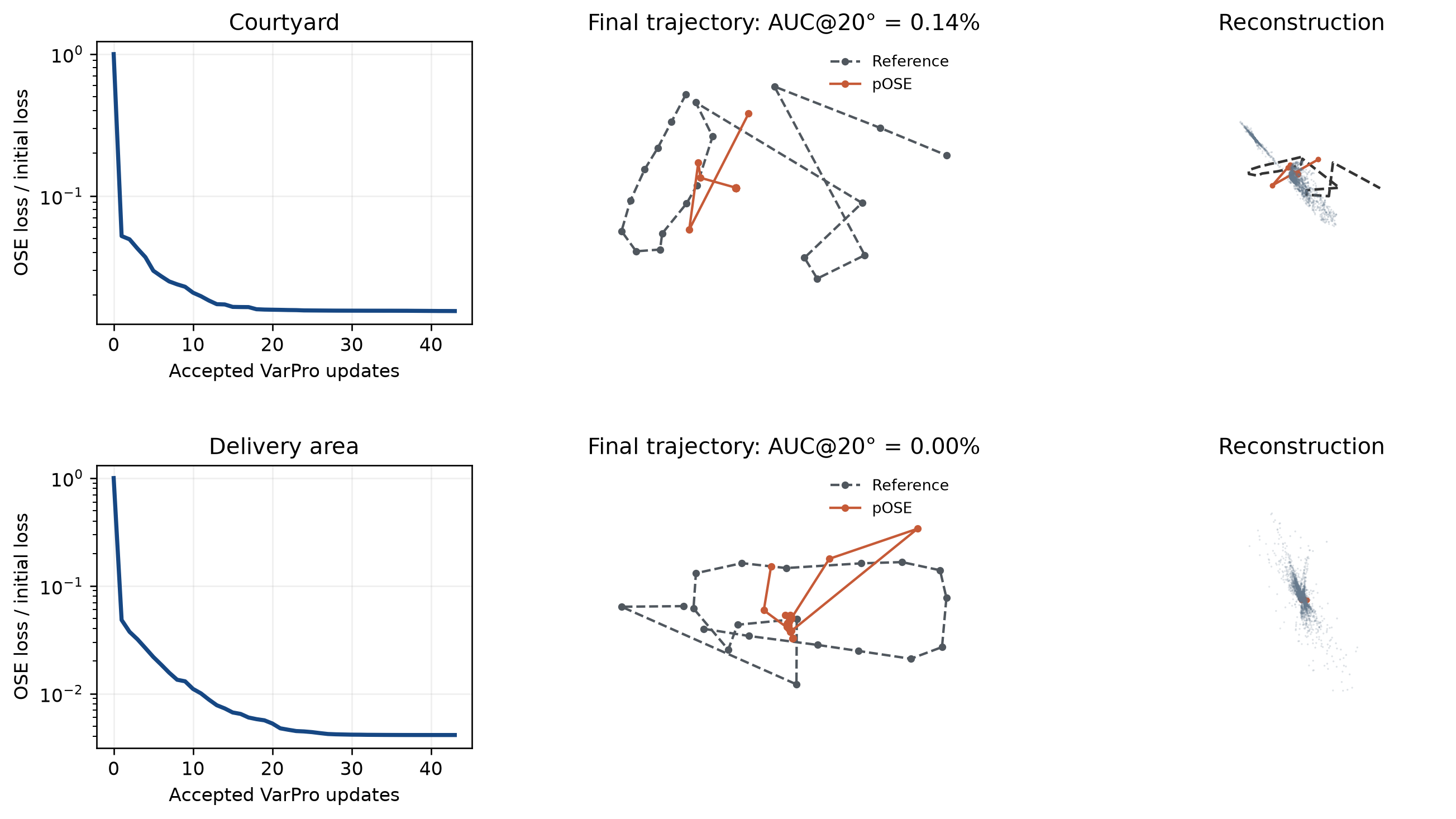}
    \caption{Optimization--reconstruction gap for the original pOSE pipeline on Courtyard (top) and Delivery area (bottom), using 20 images per scene. Left: normalized OSE loss during VarPro.
    Middle: final camera trajectories compared with the reference. Right: reconstructed landmarks and cameras. Despite decreasing surrogate losses, the final reconstructions remain inaccurate after reprojection refinement and metric upgrade, highlighting the need to reconsider how existing InitFree BA pipelines enforce metric consistency.}
    \label{fig:ose-reconstruction-gap}
\end{figure*}

\section{Convergence vs Reconstruction Gap}\label{app:gap_convergence}

Consistent with the experimental analysis of Weber et al.~\cite{weber2026initialization}, we observe a discrepancy between surrogate optimization and metric reconstruction quality in existing OSE-based formulation. As shown in \cref{fig:ose-reconstruction-gap}, the pOSE objective decreases substantially on two ETH3D scenes, yet the final camera poses and scene structure remain inaccurate after reprojection refinement and metric upgrade. These examples illustrate that successful reduction of the OSE objective alone does not ensure an accurate metric reconstruction.

\section{Formal Definition of InitFree BA}\label{app:definition}

Following the initialization-free BA literature, we use \emph{initialization-free} to mean that optimization does not require an accurate initial global reconstruction: camera poses and landmark coordinates can start from random values \cite{zach2018pose,iglesias2021radial,iglesias2023expose,olsson2025towards}. This concerns the initialization of the unknowns, rather than the absence of geometric information in the input or objective.

This distinction is consistent with previous work. The original pOSE paper \cite{zach2018pose} explicitly identifies robust matching and relative-pose verification as upstream stages of SfM. RpOSE and expOSE \cite{iglesias2021radial,iglesias2023expose} focus on reconstruction from given image measurements, leaving correspondence extraction outside their optimization problem; this does not exclude
geometric verification upstream. More explicitly, pOSE+rot \cite{olsson2025towards} incorporates relative rotation estimates into its objective, with weights derived from two-view reprojection optimization involving relative translation and local structure.

Likewise, our method uses image-derived pairwise constraints to guide optimization, but does not convert them into initial global camera poses or landmark coordinates. It is therefore initialization-free in the sense established by previous work, while explicitly incorporating geometric measurements into the objective.

\section{Effect of OSE-based optimization on low-overlap problems}\label{app:random_joint}

\begin{figure}[t]
    \centering
    \includegraphics[width=0.95\linewidth]
    {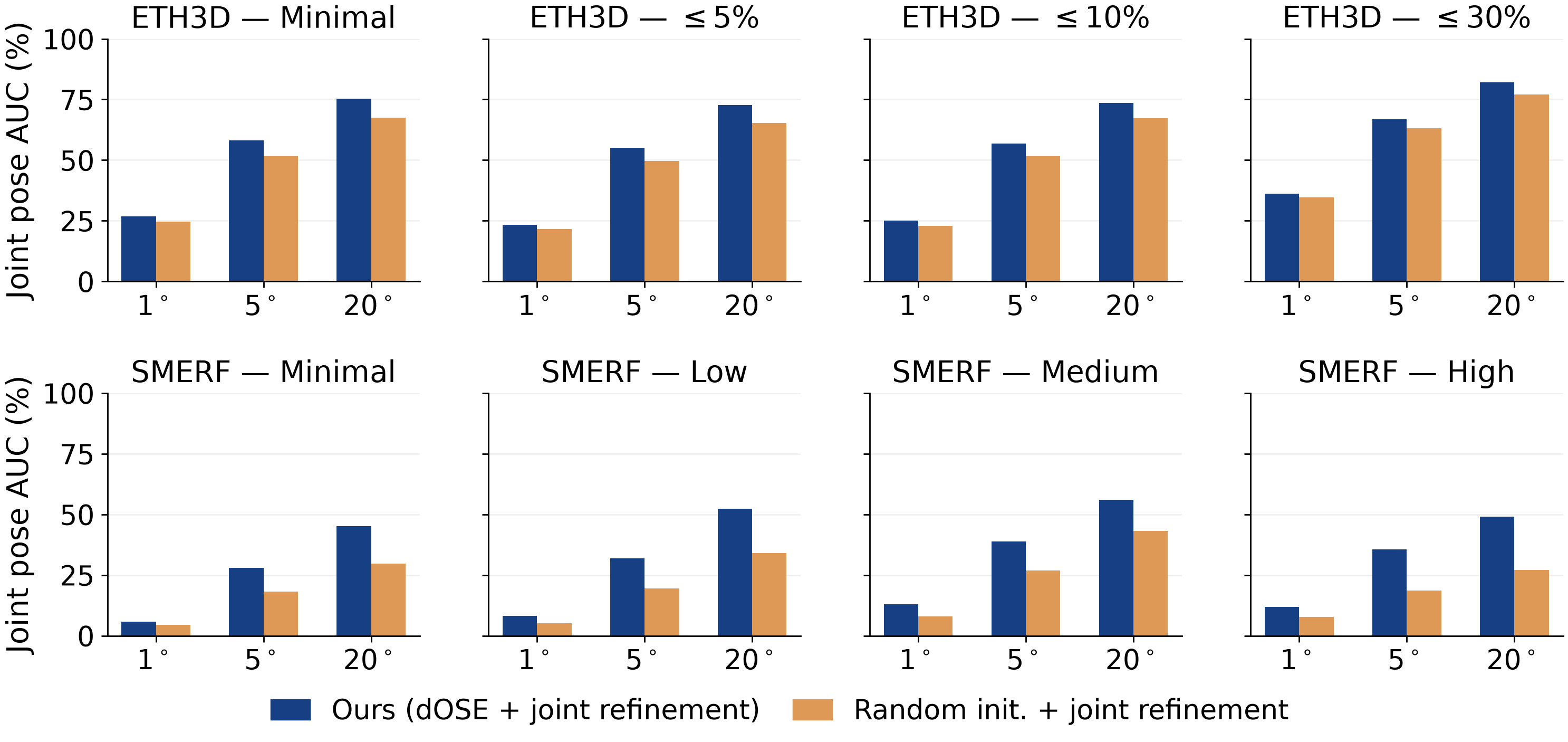}
    \caption{\textbf{Effect of OSE initialization on ETH3D low-overlap subsets (top) and SMERF subsets (bottom)} with LoMa. We compare the full pipeline (dOSE from random initialization followed by joint refinement) with joint refinement from random initialization. Bars report joint pose AUC at $1^\circ$, $5^\circ$, and $20^\circ$.}
    \label{fig:joint_vs_full}
\end{figure}

Although joint refinement from random initialization already produces meaningful reconstructions, dOSE initialization improves accuracy at all three thresholds in every low-overlap group (\cref{fig:joint_vs_full}). For instance, at AUC@$20^\circ$, the gains range from 4.8--7.9 percentage points on ETH3D and 12.9--22.0 points on SMERF. These results show that OSE-based optimization provides useful geometric guidance beyond the constraints already present in joint refinement, with particularly substantial benefits on SMERF.

\section{Effect of depth-filtering}\label{app:depth_filtering}

\begin{table}[t]
\centering
\small
\setlength{\tabcolsep}{4pt}
\caption{Effect of correspondence filtering with LoMa and MoGe-2. Entries report joint pose AUC@$1^\circ/5^\circ/20^\circ$ (\%). Only filtering of the optimized tracks differs.}
\label{tab:filtering-ablation}
\begin{tabular}{lcc}
\toprule
Track filtering & ETH3D & LaMAR \\
\midrule
Geometric only
& 32.25/60.04/75.51
& 19.35/56.24/74.57 \\
Geometric + depth
& \textbf{44.40/72.63/85.81}
& \textbf{20.23/59.61/80.99} \\
\bottomrule
\end{tabular}
\end{table}

As explained in \cref{subsec:depth_consistent_correspondences}, we complement image-based geometric verification with depth-consistent filtering. As shown in \cref{tab:filtering-ablation}, this additional filtering improves AUC@$20^\circ$ by 10.30 percentage points on ETH3D and 6.42 points on LaMAR, while also improving accuracy at tighter thresholds. This indicates that depth provides useful additional information for rejecting correspondences that survive image-based verification. Nevertheless, image-based filtering alone already yields competitive accuracy,

\section{Joint Metric Refinement}
\label{app:warm_up}

\begin{table}[t]
\centering
\scriptsize
\setlength{\tabcolsep}{3pt}
\renewcommand{\arraystretch}{1.15}
\caption{Refinement schedule ablation with LoMa and MoGe-2. Entries report joint pose AUC@1$^\circ$/5$^\circ$/20$^\circ$ (\%).}
\label{tab:refinement-schedule}
\begin{tabular}{@{}lcc@{}}
\toprule
Translation schedule & ETH3D & LaMAR \\
\midrule
Directions only
& 40.17/66.08/78.26
& 20.25/56.97/75.46 \\
Metric vectors only
& 20.54/60.59/82.75
& 14.76/56.36/78.28 \\
Metric $\rightarrow$ directions
& 44.40/72.63/85.81
& 20.23/59.61/80.99 \\
\bottomrule
\end{tabular}
\end{table}

\paragraph{Refinement schedule.}
As shown in \cref{tab:refinement-schedule}, using metric translation constraints throughout refinement improves AUC@$20^\circ$, but reduces accuracy at stricter thresholds compared to direction-only constraints. Our default approach first uses metric translations, then switches to translation directions, achieving better overall accuracy.

\paragraph{Cheirality and scale regularization.}
Let $z_{ij}=[R_iX_j+t_i]_3$ denote the reconstructed camera-space depth and $\hat d_{ij}=\max(\alpha_i d_{ij},10^{-4})$, $\tau = 0.05$.
We use the smooth cheirality penalty
\begin{equation}
E_{\mathrm{cheirality}}
=
\frac{\tau}{|\mathcal O|}
\sum_{(i,j)\in\mathcal O}
\log\left(1+\exp\left[-\frac{z_{ij}}{\tau\hat d_{ij}}\right]\right),
\end{equation}
to discourage points behind their observing cameras. Depth scales are parameterized as $\ell_i=\log\alpha_i$ and regularized by
\begin{equation}
E_{\mathrm{scale}}
=
\frac{1}{M-1}
\sum_{i=1}^{M-1}
(\ell_i-\ell_i^{\mathrm{init}})^2,
\end{equation}
where the first camera's scale is fixed, and $\ell_i^{\mathrm{init}}$ denotes the scale at the start of the current refinement stage.

\section{Implementation settings}\label{app:settings}

\paragraph{Random initialization.}
Camera rotations are sampled uniformly on $\mathrm{SO}(3)$ by drawing four independent standard Gaussian variables and normalizing the resulting quaternion. Each translation coordinate is independently sampled from a zero-mean Gaussian with standard deviation \(0.5\). The first camera pose is fixed to $(I,\mathbf{0})$ to remove the global rigid gauge, and all depth scales $\alpha_i$ are initialized at 1.0. Landmark coordinates are obtained through the closed-form solution of VarPro, for given pose parameters (\cref{eq:closed_form}). We use seed zero for the main paper. \cref{tab:seed-robustness} shows the robustness of the results across three seeds. This indicates that the aggregate results are robust to the tested random initializations.

For the joint-refinement-only experiment (\cref{subsec:ablation}), we use the same pose and depth scale distributions and additionally sample each global landmark independently from a zero-mean Gaussian with standard deviation \(0.5\). 

\begin{table}[t]
\centering
\small
\setlength{\tabcolsep}{4pt}
\caption{Seed robustness of InitFree-SfM (LoMa). Values are mean $\pm$ sample standard deviation over seeds 0, 1, and 2.}
\label{tab:seed-robustness}
\begin{tabular}{lccc}
\toprule
Dataset & AUC@$1^\circ$ & AUC@$5^\circ$ & AUC@$20^\circ$ \\
\midrule
LaMAR
& $20.44 \pm 0.29$ & $59.85 \pm 0.26$ & $80.76 \pm 0.77$ \\
ETH3D
& $44.37 \pm 0.18$ & $72.14 \pm 0.69$ & $85.46 \pm 0.34$ \\
\bottomrule
\end{tabular}
\end{table}

\paragraph{Implementation details.}
For each image pair, we compute the angular discrepancy between its depth-derived relative rotation and the rotation implied by the strongest-edge spanning tree, and pairs with discrepancy greater than \(\tau_R=15^\circ\) are rejected.
We optimize dOSE using VarPro for at most 60 iterations. We approximate the Huber depth penalty using IRLS, fixing robust weights during each VarPro elimination and updating them after accepted steps, starting after the first continuation stage.
We employ a depth-weight continuation schedule, following the general principle of progressively transitioning toward the target objective \cite{allgower2012numerical}. Specifically, the weight $\lambda_d$ is progressively increased to 5\%, 25\%, and 100\% of its target value $\lambda_d = 30$ over three 20-iteration stages. Note that we selected $\lambda_d=30$ during preliminary experiments and retained it throughout the subsequent evaluations, without dataset-specific tuning.  During the last stage, optimization terminates early when the objective decrease satisfies $E_{\mathrm{old}}-E_{\mathrm{new}} <10^{-9}\max(1,E_{\mathrm{new}})$.
%Following common practice in nonconvex optimization \cite{li2026fastmap}, we allow up to three attempts from independent random initializations, stopping at the first solution with at least 95\% positive-depth observations and a median reprojection error below 5 pixels.
The resulting reconstruction undergoes joint metric refinement using Levenberg--Marquardt algorithm and preconditioned conjugate gradients, with a 30-iteration warm-up followed by at most 120 refinement iterations. Each refinement stage terminates early if the relative parameter update falls below $10^{-8}$ or the relative objective decrease falls below $10^{-10}$. Both optimizers also terminate upon persistent step rejection when the damping parameter reaches $10^{12}$.

\paragraph{Dataset selection.}
For ETH3D \cite{schops2017multi}, we evaluate 25 scenes using up to 20 images per scene, selected at uniformly spaced indices in the filename-sorted image list; scenes with fewer than 20 images use all available images. For LaMAR \cite{sarlin2022lamar}, we construct 45 subsets of 20 RGB phone images, with 15 subsets from distinct captures in each of CAB, HGE, and LIN. Using a fixed random seed, we select one temporally contiguous window per capture. Reference poses are used only during subset construction to exclude abrupt motion and near-static windows: consecutive frames must be within 3\,m, $45^\circ$, and 5\,s, and each window must include at least 0.5\,m displacement from its first camera. All methods receive the same frozen image selections. For low-overlap ETH3D and SMERF, we follow the released MP-SfM benchmark subsets and overlap groups \cite{pataki2025mpsfm}.

\paragraph{Evaluation metric.}
Following current practice, we define joint pose error as the maximum of the relative rotation error and the sign-invariant translation-direction error for each camera pair. We report the area under the recall curve (AUC) up to $1^\circ$, $5^\circ$, and $20^\circ$, normalized to percentages.

\paragraph{Baseline configurations.}
For Depth Anything 3, we use DA3-Large-1.1; for MP-SfM, we use the Metric3Dv2 Giant preset for depth and normal estimation. In the main ETH3D and LaMAR comparisons, LoMa-based methods use LoMa-B with 2048 keypoints per image and a confidence threshold of 0.10. In the low-overlap comparisons, our method and GLOMAP use LoMa-B with 4096 keypoints. For the RoMa-v2 comparisons, we use the precise preset and sample 1024 correspondences per image pair. Geometric methods use dataset calibration in the calibrated comparisons, while feed-forward methods and VGGSfM estimate their own calibration. Each method retains its own verification and reconstruction pipeline.

\begin{table}[t]
\centering
\small
\setlength{\tabcolsep}{3pt}
\caption{Relative rotation penalties and depth-scale adaptation during VarPro. Joint pose AUC@$1^\circ/5^\circ/20^\circ$ (\%). All variants use LoMa and MoGe-2 depth.}
\label{tab:dose-rot}
\begin{tabular}{lccc}
\toprule
Method & Variable $\alpha$ & ETH3D & LaMAR \\
       & in VarPro & & \\
\midrule
Ours (full)
& No
& 44.40/72.63/85.81
& 20.23/59.61/\textbf{80.99} \\
dOSE
& Yes
& 44.35/72.43/85.01
& 20.06/58.38/79.25 \\
dOSE+rot
& Yes
& \textbf{44.73/74.27/87.58}
& \textbf{20.51/60.07}/80.98 \\
\bottomrule
\end{tabular}
\end{table}

\section{VarPro algorithm}\label{app:varpro}

\paragraph{Variable Projection.}
Let
\begin{equation}
    u = \{R_i,t_i\}_{i=1}^{M-1},
    \qquad
    v = \{X_j\}_{j=1}^{N},
\end{equation}
where the first camera is fixed to remove the global gauge. The resulting least-squares problem can be written in separable form as
\begin{equation}
    \min_{u,v} \varepsilon(u,v) =
    \left\lVert G(u)v-z(u)\right\rVert_2^2,
    \qquad
    (R_i,t_i)\in SE(3).
\end{equation}

For fixed camera poses $u$, the optimal landmarks can be eliminated in closed form,
\begin{align}\label{eq:closed_form}
    v^{*}(u)
    &= \argmin_v
    \left\lVert G(u)v-z(u)\right\rVert_2^2 \\
    &= G(u)^{\dagger}z(u).
\end{align}
The optimization can therefore be performed solely over the camera poses, while the optimal scene structure is implicitly updated at each iteration. This substantially reduces the dimension of the nonlinear optimization problem.

Following \cite{hong2017revisiting}, we use the Ruhe--Wedin (RW2) approximation \cite{kaufman1975variable} to obtain the Jacobian of the projected residual. Let $J_u$ and $J_v$ denote the Jacobians with respect to the pose and landmark variables, respectively, evaluated at $(u,v^{*}(u))$. The projected pose Jacobian is
\begin{equation}
    J_u^{*}
    =
    (I-J_vJ_v^{\dagger})J_u,
\end{equation}
where $(I-J_vJ_v^{\dagger})$ projects out residual variations that can be explained by an update of the scene structure. The resulting Levenberg--Marquardt pose update is obtained from
\begin{equation}\label{eq:delta_u}
    \left(
        J_u^{\top}(I-J_vJ_v^{\dagger})J_u
        +\lambda I
    \right)\Delta u
    =
    -{J_u^{*}}^{\top}\varepsilon.
\end{equation}
Thus, each iteration first eliminates the scene structure and then updates only the camera poses.

For small-scale problems, we solve \Cref{eq:delta_u} directly using a Cholesky decomposition. For larger problems, explicitly forming and factorizing the projected system becomes expensive. We therefore use the inverse-expansion strategy of Weber et al. \cite{weber2023power,weber2024power}, which approximates the inverse of the projected system through a
truncated power series:
\begin{align}
    \Delta u =
    - \sum_{i=0}^{m}
    \Big(
        (J_u^{\top}J_u&+\lambda I)^{-1} J_u^{\top}J_vJ_v^{\dagger}J_u
    \Big)^i \\
    &\cdot (J_u^{\top}J_u+\lambda I)^{-1}
    {J_u^{*}}^{\top}\varepsilon.
\end{align}
The resulting increment $\Delta u$ is finally applied on the $SE(3)$ manifold, with rotations updated through the exponential map
and translations updated additively.

%\section{Runtime}

\section{More sophisticated OSE}\label{app:sophisticated}

%\paragraph{Relative rotation constraints during dOSE optimization.}
Inspired by pOSE+rot \cite{olsson2025towards}, we investigate augmenting dOSE with relative rotation penalties during VarPro. The penalties use information matrices derived from linearized two-view reprojection optimization, accounting for the local translation and landmark parameters. Unlike our default pipeline, which introduces pairwise geometric constraints during joint refinement, this variant also incorporates rotation information in the initial dOSE stage. In these experiments, depth scales $\alpha_i$ are variable during both VarPro and refinement. As shown in \cref{tab:dose-rot}, adding rotation penalties improves accuracy over dOSE with variable scales on both datasets. Compared with our default pipeline, it improves ETH3D accuracy, while yielding similar results on LaMAR. These results suggest that further improvements to the OSE formulation remain a promising direction for future research.